\documentclass{article}

\PassOptionsToPackage{numbers, compress}{natbib}
\usepackage[final]{neurips_2024}

\usepackage[utf8]{inputenc} 
\usepackage[T1]{fontenc}    
\usepackage{hyperref}       
\usepackage{url}            
\usepackage{booktabs}       
\usepackage{amsfonts}       
\usepackage{nicefrac}       
\usepackage{microtype}      
\usepackage{xcolor}         
\usepackage{amsmath}
\usepackage{multirow, array}
\usepackage{graphicx}
\usepackage{amsmath}
\usepackage{algorithm}
\usepackage{algpseudocode}

\title{Procedure-Aware Surgical Video-language Pretraining with Hierarchical Knowledge Augmentation}

\author{%
  Kun Yuan${^{1, 2, 3}}$ \hspace{4mm} Vinkle Srivastav$^{1, 2}$\hspace{5mm} Nassir Navab$^3$\hspace{5mm} Nicolas Padoy$^{1, 2}$\hspace{4mm} \\[.15in]
  $^1$University of Strasbourg, CNRS, INSERM, ICube, UMR7357, Strasbourg, France \\
  $^2$IHU Strasbourg, Strasbourg, France \\ 
  $^3$CAMP, Technische Universit\"at M\"unchen, Munich, Germany
}

\begin{document}

\maketitle

\begin{abstract}
Surgical video-language pretraining (VLP) faces unique challenges due to the knowledge domain gap and the scarcity of multi-modal data. This study aims to bridge the gap by addressing issues regarding textual information loss in surgical lecture videos and the spatial-temporal challenges of surgical VLP. We propose a hierarchical knowledge augmentation approach and a novel Procedure-Encoded Surgical Knowledge-Augmented Video-Language Pretraining (PeskaVLP) framework to tackle these issues. The knowledge augmentation uses large language models (LLM) for refining and enriching surgical concepts, thus providing comprehensive language supervision and reducing the risk of overfitting. PeskaVLP combines language supervision with visual self-supervision, constructing hard negative samples and employing a Dynamic Time Warping (DTW) based loss function to effectively comprehend the cross-modal procedural alignment. Extensive experiments on multiple public surgical scene understanding and cross-modal retrieval datasets show that our proposed method significantly improves zero-shot transferring performance and offers a generalist visual representation for further advancements in surgical scene understanding. The code is available at https://github.com/CAMMA-public/SurgVLP
\end{abstract}

\section{Introduction}

The recent advancements in multi-modal representation learning, particularly with the introduction of CLIP~\citep{radford2021learning}, have led to the development of models capable of understanding a wide range of visual concepts using natural language supervision~\citep{li2023unmasked,luddecke2022image}. The expressive natural language has allowed these models to shift from task-specific to more generalist applications~\citep{ni2022expanding,zou2023generalized,zou2024segment}. The learned representations of these models are robust, facilitating effective performance across diverse visual tasks without the need for task-specific fine-tuning~\citep{wang2018reconstruction,zhou2018end}. However, despite the impressive progress made by these models in the general computer vision domain, the effectiveness of these methods in domain-specific settings remains uncertain. 

This concern is particularly relevant to the field of Surgical Data Science (SDS), an emerging interdisciplinary domain that utilizes deep learning and computer vision techniques to analyze surgical data~\citep{maier2017surgical,maier2022surgical,yuan2021surgical}. A key component of SDS is the analysis of intraoperative surgical videos captured through endoscopes or laparoscopes. Analyzing these videos presents several unique challenges compared to the general computer vision datasets. Unlike general computer vision datasets~\citep{miech2019howto100m,radford2021learning,bain2021frozen}, surgical videos can last several hours and capture complex and fine-grained activities within a narrow field of view. This requires development of computational approaches to decompose and model the surgical procedures at multiple hierarchical levels, including the entire procedure~\citep{kannan2019future}, phases~\citep{twinanda2016endonet,funke2023tunes}, steps~\citep{ramesh2021multi,lavanchy2023challenges}, atomic actions~\citep{ayobi2024pixel,bawa2021saras}, and action triplets~\citep{nwoye2021rendezvous,sharma2023surgical}. Moreover, surgical language involves specialized vocabulary, and annotating videos requires clinical expertise, limiting dataset scalability. Consequently, current deep learning applications are restricted to single-centric, fully-supervised, and task-specific approaches~\citep{twinanda2016endonet,wang2022autolaparo,lavanchy2023challenges,nwoye2021rendezvous,yuan2021surgical,ayobi2024pixel,rivoir2020rethinking}.

To bridge the gap, recent efforts have focused on creating surgical video-text pretraining datasets by curating surgical lecture videos from online e-learning platforms and pairing them with transcribed narrations using audio speech recognition (ASR) methods. Subsequently, a CLIP-style model~\citep{yuan2023learning} is trained contrastively to match the video clips to their corresponding textual descriptions. Building on this, the HecVL approach introduces the use of hierarchical texts, which include phase-level keystep descriptions and video-level summaries that provide hierarchical goals of the surgical procedure~\citep{yuan2024hecvl}. However, challenges persist due to the smaller size of the surgical video-language pretraining dataset, noisy transcribed narrations, limited variability in phase-level descriptions, and strong temporal dependencies in surgical procedures, where actions and keysteps occur in a specific routine order. These issues hinder the accurate learning of multi-modal surgical representations.

\begin{figure*}
\begin{center}
\includegraphics[width=0.98\linewidth]{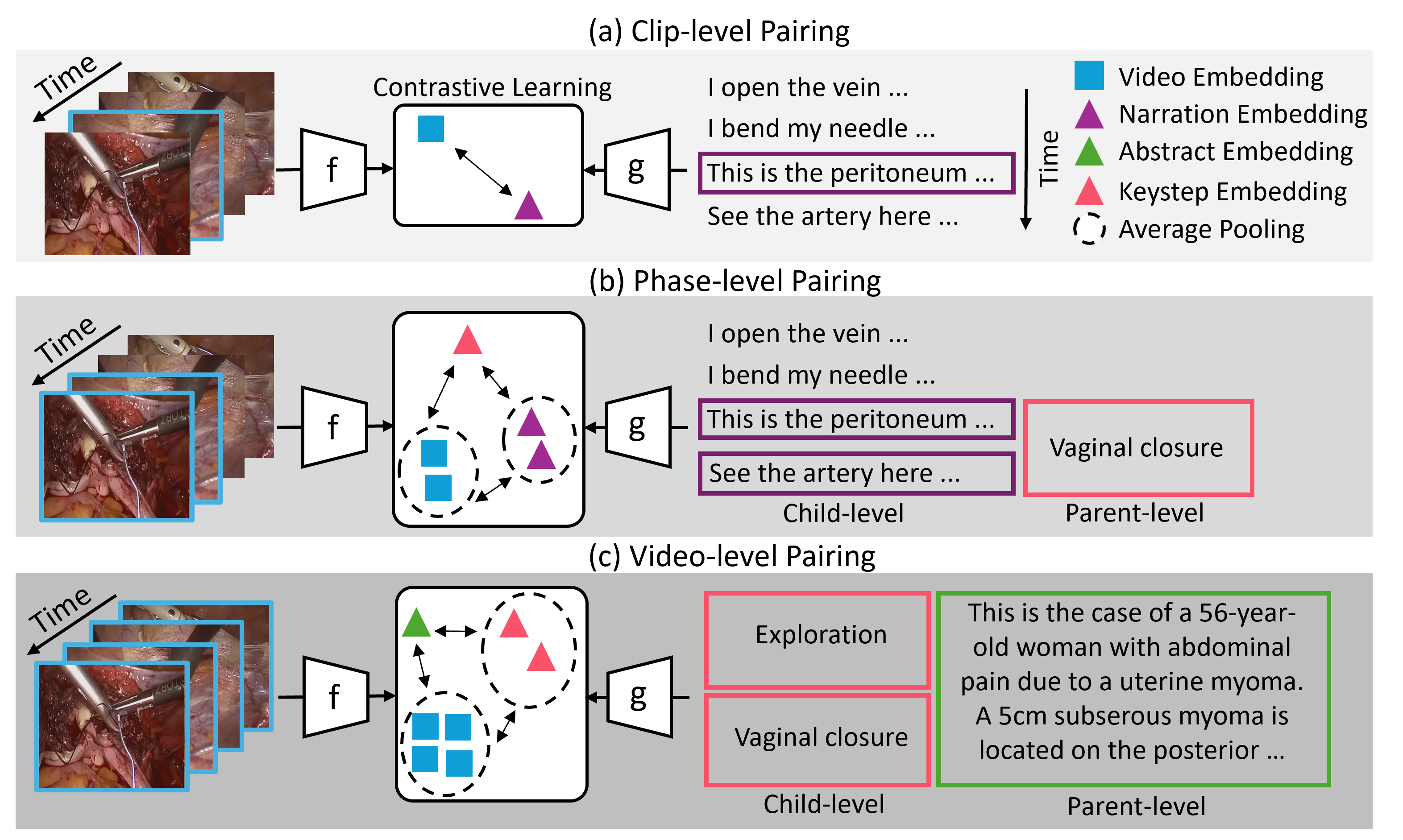}
\end{center}
\caption{\footnotesize{Illustration of video-language pretraining with hierarchical video-text pairs. At phase- and video-level, one parent-level text is paired to multiple child-level texts. }}\label{fig:overview}
\end{figure*}

To address these challenges, we propose \textbf{P}rocedure-\textbf{E}ncoded \textbf{S}urgical \textbf{K}nowledge-\textbf{A}ugmented \textbf{V}ideo-\textbf{L}anguage \textbf{P}retraining (PeskaVLP), which boosts data efficacy and tackles the spatial-temporal challenges inherent in surgical procedures from two perspectives. First, we introduce hierarchical knowledge augmentation to mitigate the problem of textual information loss in surgical video-language pretraining datasets. We argue that the internal knowledge of LLMs serves as a valuable surgical knowledge base, enriching and correcting text descriptions while preserving the original key concepts and meanings. Therefore, We utilize the large language model (LLM) prompted with different behaviors as external knowledge base to correct, explain, or summarize the hierarchical texts in the surgical video-language pretraining dataset, thus providing diverse and better language supervision for multi-modal pretraining. Additionally, it reduces the risk of overfitting by preventing the text encoder from repeatedly encountering the same keystep texts in each epoch.

From the pretraining objective perspective, we perform the hierarchical video-language pretraining, as shown in Fig.~\ref{fig:overview}, with a novel hierarchy-specific loss, $LecNCE$. Specifically, we combine language supervision with visual self-supervision at the clip-level pretraining to introduce additional supervision signals within vision modality, making the pretraining efficient with a small surgical dataset~\citep{yuan2023learning}. At phase- and video-level pretraining, we construct hard negative samples by reversing the order of texts, followed by a Dynamic Time Warping (DTW) based loss function to learn the temporal alignment between video frames and texts, thus facilitating the understanding of cross-modal procedural alignment during pretraining.

We summarize our contributions as follows: First, we propose an LLM-based knowledge augmentation to handle surgery-specific textual information loss in the dataset, providing more densely interconnected natural language supervision from surgical lecture videos. Second, our proposed hierarchical video-language pretraining method enforces the understanding of the spatial-temporal characteristics of surgical lecture videos at different hierarchical levels. The pretrained PeskaVLP demonstrates state-of-the-art transferability and visual representation to different surgical scene understanding downstream datasets~\citep{twinanda2016endonet,wang2022autolaparo,lavanchy2023challenges}, across types of surgical procedures and clinical centers. It also shows strong multi-modal alignment ability through the cross-modal retrieval task at multiple hierarchical levels. 

\section{Related Works}
\noindent\textbf{Surgical Video-Language Pretraining:}
many works have demonstrated the effectiveness of learning visual representations from the natural language supervision of corresponding text~\citep{bain2021frozen,xu2021videoclip,yuan2021multimodal,lin2022egocentric,miech2020end,luo2020univl,li2023unmasked}. These methods conduct contrastive learning~\citep{oord2018representation} to match the video clips (or images) with their corresponding narrations (or captions). Similarly in the medical field, recent works have started to curate large-scale multi-modal data through hospital-sourced chest radiological reports~\citep{johnson2019mimic,chen2024chexagent} and online platforms~\citep{yuan2023learning,ikezogwo2024quilt,huang2023visual}, e.g., YouTube and Twitter, to perform vision-language pretraining. However, these works encounter the sample efficiency issue when handling the smaller surgical video-language pretraining dataset (SVL)~\citep{yuan2023learning}. Recent works improve the data efficacy and zero-shot performance of CLIP-style models~\citep{mu2022slip,li2021supervision,huang2021gloria}. However, they do not capture procedural dependency from the long-form surgical videos beyond the video clip and text matching. Hierarchical pretraining methods~\citep{ashutosh2023hiervl,zhang2018cross,yuan2024hecvl} propose to pair video clips of different durations to different hierarchical levels of texts, covering both short- and long-term understanding. Paprika~\citep{zhou2023procedure} builds a procedural knowledge graph and elicits the knowledge node during the video-language pretraining process.

\noindent\textbf{Textual Augmentation with Knowledge Base:} the success of vision-language pretraining is highly dependent on the quality and quantity of available multi-modal data. Recent research~\citep{li2024scaling} shows that a smaller high-quality dataset can outperform a larger low-quality dataset. Common practices improve the quality by textual augmentation, including EDA~\citep{li2021supervision}, masked token modeling~\citep{sun2019videobert}, and captioning loss~\citep{yu2022coca}. Recent studies have used synthesized captions from captioning models to achieve notable improvements~\citep{li2022blip,li2023blip,rotstein2024fusecap}. However, they show scalability deficiency and world knowledge loss in models trained with synthetic captions~\citep{yu2023capsfusion}, which their initial benchmark success has largely obscured. To inject the knowledge, K-Lite~\citep{shen2022k} enriches the texts with WordNet~\citep{fellbaum1998wordnet} and Wiktionary~\citep{meyer2012wiktionary} knowledge base. Merlot~\citep{zellers2021merlot} learns script knowledge representations from millions of YouTube videos, however, a knowledge domain gap exists when applying this to the surgical field. The recent advent of self-supervised large language models like GPT4~\citep{achiam2023gpt} and Llama series~\citep{touvron2023llama} have been a game-changer, as they encode rich domain-specific knowledge, e.g., clinical knowledge~\citep{singhal2023large}, motivating LaCLIP~\citep{fan2024improving} to augment textual inputs through the LLM rewrites.
\section{Approach}

\subsection{Dataset and Contrastive Learning}
Learning joint video and language embedding space requires a large-scale video-language dataset, however, such datasets are expensive and time-consuming to create in the surgical field. Therefore, the first surgical video-language pretraining dataset, i.e., SVL~\citep{yuan2023learning}, is proposed by obtaining around a thousand surgical lecture videos from surgical education platforms. SVL collects $\sim$$300$ hours of lecture videos accompanied by narration texts obtained using Audio Speech Recognition (ASR) methods, providing $\sim$$26$k video clip-narration pairs for contrastive video-language pretraining. Specifically, short video clips $x_c$ and their corresponding narration texts $y_n$ are treated as positive pairs $\mathcal{P}^n$, and the unpaired ones are treated as negative pairs $\mathcal{N}^n$. Then, the contrastive training loss $InfoNCE$~\citep{oord2018representation} can be formulated as follows:
\begin{equation}
L_{InfoNCE} = \max_{f,g}\sum_{i=1}^B\log\hspace*{-1mm}\left(\frac{\sum\limits_{(x_c, y_n)\in\mathcal{P}_i^n}e^{ f(x_c)^\top g(y_n)}}{\hspace*{-1mm}\sum\limits_{(x_c, y_n)\in\mathcal{P}_i^n}\hspace*{-2mm}e^{ f(x_c)^\top g(y_n)}+\hspace*{-5mm}\sum\limits_{(x_c',y_n')\sim\mathcal{N}_i^n}\hspace*{-4mm}e^{f(x_c')^\top g(y_n')}}\right)\hspace*{-1mm}
\label{loss:narration_clip}
\end{equation} where $B$ represents the batch size. The $f$ and $g$ are visual and textual encoders that generate embedding vectors for videos and texts, respectively. This loss function aligns two modalities by increasing the cosine similarity between paired videos and texts and decreasing the unpaired ones, as shown in Fig.~\ref{fig:overview} (a). Despite reaching an impressive data scale, the imprecision of the ASR system and the scarcity of surgical lecture videos limit the natural language supervision from SVL. Therefore, HecVL~\citep{yuan2024hecvl} proposes to incorporate hierarchical language supervision by extracting additional phase-level keystep and video-level abstract texts from lecture videos' metadata, as shown in Fig.~\ref{fig:overview} (b) and (c). In this work, we use this hierarchical video-language pretraining dataset and perform hierarchical knowledge augmentation to improve the textual quality.

\subsection{Hierarchical Knowledge Augmentation}

\begin{figure*}
\begin{center}
\includegraphics[width=0.98\linewidth]{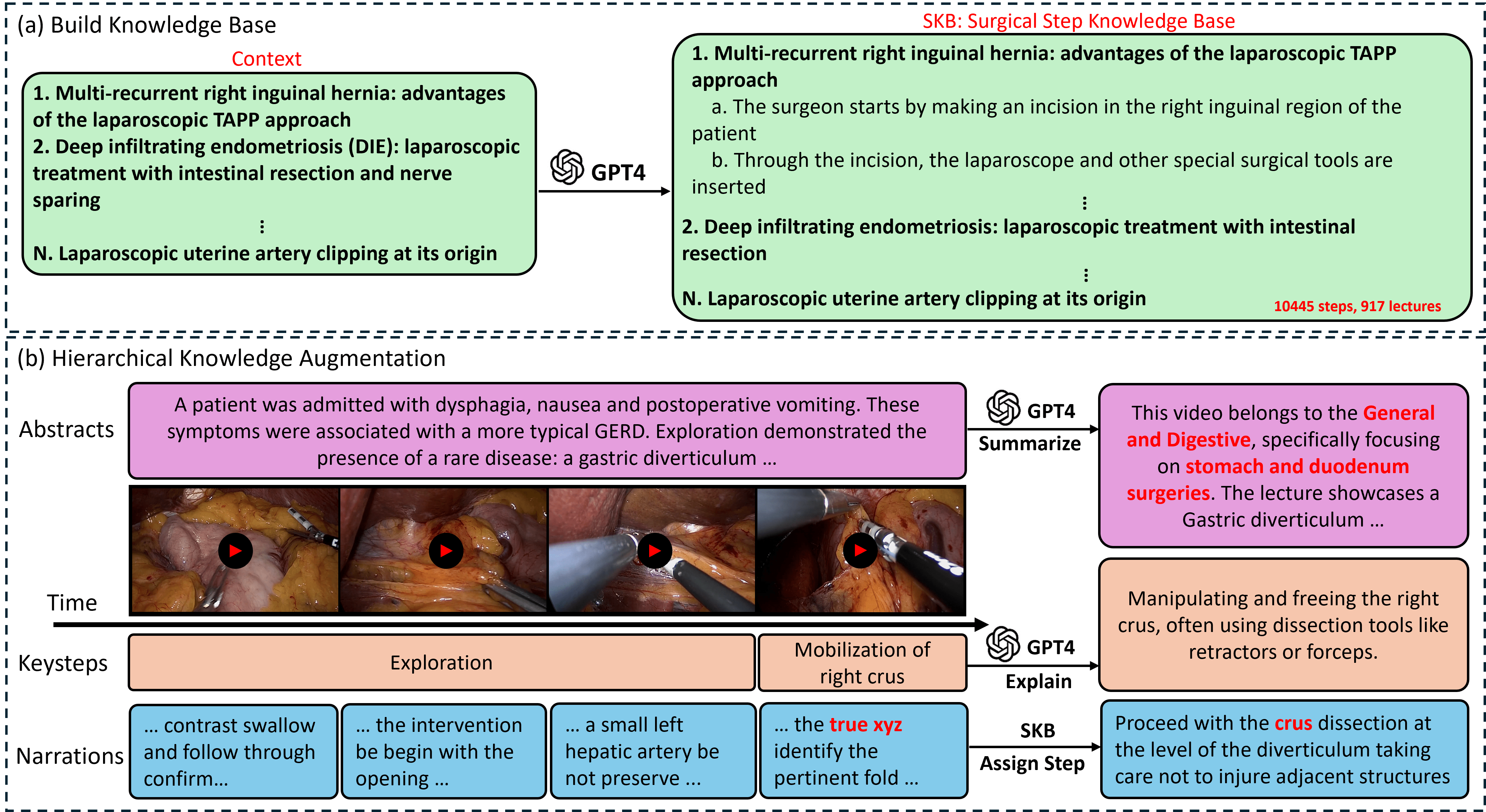}
\end{center}
\caption{\footnotesize{Hierarchical Knowledge augmentation for hierarchical texts. (a) the process of building a surgical step knowledge base. (b) the process of improving hierarchical textual quality based on LLM.}}\label{fig:rewrite}
\end{figure*}

Quality of language supervision matters~\citep{abbas2023semdedup,li2021supervision,li2024inverse} especially when the surgical video-language dataset is not ``big'' enough, e.g., millions of multi-modal samples used in ~\citep{radford2021learning,miech2019howto100m}, to sufficiently cover the visual-linguistic concepts. In this work, we find that the texts suffer from different types of degradation at different hierarchies, failing to provide accurate and broad concepts for pretraining. Specifically, as shown in Fig.~\ref{fig:rewrite}, narration texts are mostly sentence fragments and easily affected by misspelling errors, therefore altering the original key concepts. The keystep texts are mostly short and abstract, resulting in a narrow set of linguistic concepts that could show poor transferability to the downstream datasets, which usually come with a different set of concepts~\citep{shen2022k,geirhos2018imagenet}. The abstract texts sometimes include redundant and useless information, such as author and citation information. 

To address the above hierarchy-specific textual degradation, we propose a hierarchical knowledge augmentation to correct/explain/summarize the narration/keystep/abstract texts, respectively, by eliciting LLM's encoded surgical knowledge~\citep{singhal2023large}. For each hierarchy, we manually design the system prompt and several input-output examples for LLM. Thus, we obtain hierarchical LLM assistants with different behaviors of using internal surgical knowledge to augment the texts:

\textbf{Narration.} We ask the LLM to behave as a ``recipe'' to come up with a list of sequential steps that complete the given surgery. For each lecture video, we feed its title as input and obtain the list of pseudo steps, as shown in Fig.~\ref{fig:rewrite} (a), building a surgical step knowledge base. Then, we assign these pseudo steps to narration texts based on textual similarity. This implicitly corrects the typos in transcribed narrations and augments the textual input based on the LLM's surgical knowledge. 
\textbf{Keystep.} As shown in Fig.~\ref{fig:rewrite} (b), we ask the LLM to behave like a ``dictionary'' to explain the meaning of the keystep. Specifically, the LLM assistant expands the given keystep into a description of the main surgical events, anatomies, and instruments involved. This enlarges the textual semantic information of each keystep and provides more expressive language supervision for pretraining.
\textbf{Abstract.} As shown in Fig.~\ref{fig:rewrite} (b), we ask the LLM to behave like a ``summarizer'' that captures the key concepts of the given abstract texts, e.g., surgical type, anatomies, and so on. This reduces the length of the textual inputs while maintaining the main concepts of the abstract paragraph. In the following experiment, we randomly input the original or augmented texts for video-language pretraining. Check Appendix H for examples of pre- and post-augmented texts.

\subsection{Procedure-aware Surgical Video-language Pretraining}

\begin{figure*}
\begin{center}
\includegraphics[width=0.98\linewidth]{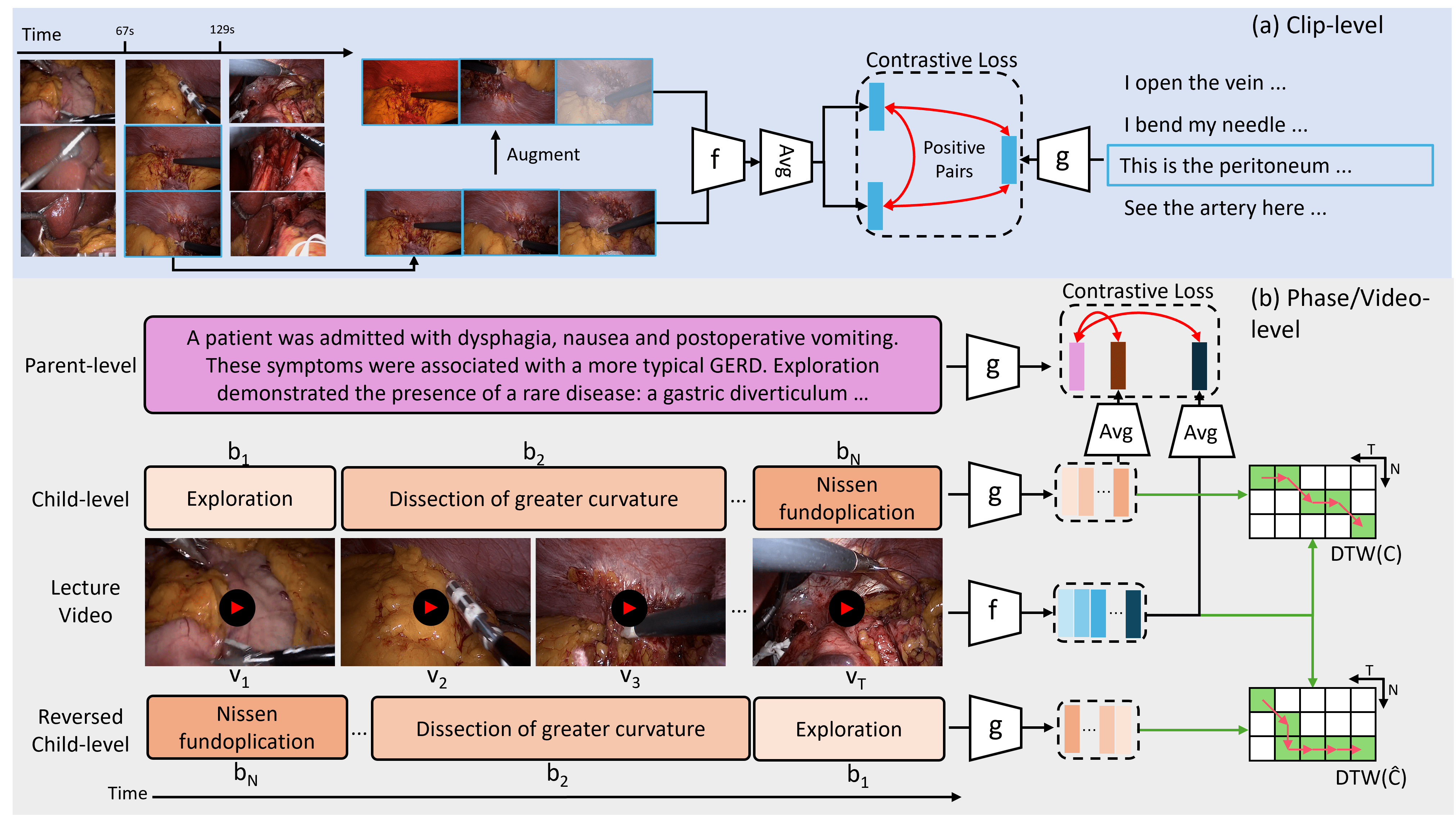}
\end{center}
\caption{\footnotesize{The pretraining pipeline of 
different hierarchies. We combine language supervision and visual self-supervision at clip-level pretraining. We conduct the procedure-aware contrastive learning at phase/video-level pretraining.}}\label{fig:pipeline}
\end{figure*}

We introduce PeskaVLP, a procedure-aware pretraining framework for the above surgical knowledge-augmented video-language dataset. We emphasize devising a pretraining objective $LecNCE$ for the hierarchical video-text pairs. For clip-level pretraining, $LecNCE_{clip}$ combines language supervision with visual self-supervision to improve data efficiency and boost the scene understanding on visually similar laparoscopic images. $LecNCE_{phase/video}$ considers the procedure awareness during the coarser-level pretraining, through a DTW-based contrastive regularization objective with temporally reversed text sequences as negative samples. We apply the dual-encoder as our model architecture.

\subsubsection{Clip-level Pretraining}

\textbf{Language Supervision.} The common pretraining objective for dual-encoder model is $InfoNCE$~\citep{oord2018representation}, as denoted in Eq.~\ref{loss:narration_clip}, where matched video text pairs are treated as positive while all other pairwise combinations in the batch are regarded as negative. In this work, we also apply $InfoNCE$ to maximize the similarity between short-term video clips and their corresponding narration texts at the clip level, denoted as $L^{vl}_{clip}$. However, this simple objective is data hungry and sensitive to the weakly aligned noisy video-text pairs from small-scale surgical video-language datasets, such as SVL~\citep{yuan2023learning}.

\textbf{Visual Self-supervision.} To address that, our PeskaVLP introduces an additional supervisory signal from visual self-supervision to complement noisy language supervision. Specifically, we explore the widespread supervision within visual modality to learn generic visual representation. We adopt the simple yet effective SimSiam~\citep{chen2021exploring} strategy, whose objective is to maximize the similarity between two augmented views. As shown in Fig.~\ref{fig:pipeline} (a), during the pretraining, we apply random distortion on the frames of video clips and generate two augmented embedding vectors for one video clip. We then apply $InfoNCE$ to maximize the similarity of these two augmented embeddings by treating them as positive pairs, denoted as $L^{vv}_{clip}$. This additional supervisory can learn visual features more efficiently and is robust to the distortion of surgical scene images. Finally, the $LecNCE$ loss for clip-level pretraining is the sum of these two losses, denoted as $LecNCE_{clip} = L^{vl}_{clip} + L^{vv}_{clip}$.

\subsubsection{Phase-/Video-level Pretraining}
The surgical video-language pretraining presents a unique procedural challenge compared to the existing video-language methods~\citep{grauman2022ego4d,miech2019howto100m,radford2021learning,xue2024learning,sermanet2018time}. The surgical actions and events occur in a certain order to follow the routine to complete the surgical phase and surgery, e.g., ``hook dissecting cystic duct'' should happen before ``clipper cutting cystic duct'' in the ``clipping cutting'' phase of cholecystectomy surgery. However, prior contrastive learning objectives~\citep{miech2020end,radford2021learning,grauman2022ego4d} omit this temporal dependency and limit the understanding of procedural knowledge in surgical lecture videos. 

Our proposed $LecNCE$ training objective enables procedural understanding in phase- and video-level pretraining by considering the cross-modal temporal alignment between video frames and text sequence. Specifically, hierarchical texts can form the parent-child correspondence, i.e., abstract (parent-level) and keystep (child-level) texts, keystep (parent-level) and narration (child-level) texts. As shown in Fig.~\ref{fig:pipeline} (b), each parent-level text $A$ is paired with a video segment $V = \{v_1,...v_T\}$, where the $T$ is the number of frames of the video segment. $A$ is also paired with a child-level text sequence $B = \{b_1,...b_N\}$, where $N$ is the length of this sequence. Then, we build the cost matrix $C \in R^{T \times N}$ between video frames and child-level text sequence based on their embeddings, with each element $c_{i,j}$ computed by a distance function $D$. We adopt the same distance function from~\citep{hadji2021representation}:
\begin{equation}
    c_{i,j} = \mathcal D(v_i, b_j) = -\log \frac{\exp(\tilde{\mathbf{v}}_i^\top \tilde{\mathbf{b}}_j / \beta)} {\sum_{k=1}^N \exp(\tilde{\mathbf{v}}_i^\top \tilde{\mathbf{b}}_k / \beta)}, \quad \tilde{\mathbf{v}}_i = f(v_i) / \| f(v_i) \|_2 \quad \tilde{\mathbf{b}}_i = g(b_i) / \| g(b_i) \|_2
    \label{loss:distance_dunction}
\end{equation}
Using this cost matrix $C$, we apply Dynamic Time Warping (DTW) to find the minimum cross-modal cost path that aligns the video frames to the text sequence, denoted as $DTW(C)$. We then make a reasonable assumption that the global semantics of the text sequence and its reversed version are distinct. Therefore, aligning the video frames to the text sequence should be easier, i.e., incur a lower alignment cost compared to aligning the same video frames when the text sequence is played in reverse. Following this assumption, we temporally reverse the child-level texts into $\hat{B} = \{b_n,...b_1\}$ and build the cost matrix $\hat{C}$ between $V$ and $\hat{B}$, computing the minimum alignment cost $DTW(\hat{C})$. We then devise a DTW-based contrastive regularization using hinge loss as follows: 
\begin{equation}
L_{dtw} = max(DTW(C) - DTW(\hat{C})), \phi)
\label{loss:dtw}
\end{equation}
where $\phi$ is the margin between positive and negative samples. This imposed regularization can support fine-grained multi-modal representation learning from weakly paired video frames and texts via temporal alignment. 
Unlike Paprika~\citep{zhou2023procedure}, which relies on a pretrained model~\citep{miech2020end}, our phase-/video-level pretraining provides a direct, lightweight, and more adaptable methodology to unseen surgical domains. We do not require the adaption from any existing models, improving the generalization capability. Also, our pretraining process is procedure-aware in itself rather than modifying the representation in a second step, streamlining the process and increasing efficiency.
We also apply the $InfoNCE$ loss to maximize the similarity  between the paired parent-level text, video segment, and child-level texts, denoted as $L_{infonce}$. Note that the $L_{infonce}$ follows the same pipeline as in Fig.~\ref{fig:overview} (b) and (c). Finally, we achieve the loss $LecNCE$ for phase- or video-level pretraining as $LecNCE_{phase/video} = L_{infonce} + \lambda L_{dtw}$, where $\lambda$ is the hyper-parameter to scale two losses. Please refer to Appendix D for more details about dynamic time warping. 
Finally, we train the model in an alternating way, using the proposed hierarchical levels of learning objectives. We only train one set of visual and textual encoders for all three levels, ensuring the encoders are optimized for capturing both short-term and long-term semantics. We alternatively train with $25$ batches of clip-level samples, followed by $15$ and $115$ batches of phase- and video-level samples.

\section{Experiments}

\textbf{Datasets.}
Our pretraining is conducted on the videos of SVL~\citep{yuan2023learning} dataset. The pertaining dataset includes hierarchical textual annotations from the metadata of the videos~\citep{yuan2024hecvl}. We evaluate our model on $3$ publicly available surgical phase recognition downstream datasets, i.e., Cholec80~\citep{twinanda2016endonet} (cholecystectomy) from Strasbourg center, AutoLaparo~\citep{wang2022autolaparo} (hysterectomy) from HongKong hospital, MultiBypass140~\citep{lavanchy2023challenges} (gastric bypass) from both Strasbourg (StrasBypass70) and Bern (BernBypass70) centers. These datasets contain untrimmed surgical workflows with frame-wise phase labels. We also evaluate pretrained model on the cross-modal retrieval task in multiple hierarchical levels with the holdout videos in SVL-Retrieval~\citep{yuan2023learning}. Check Appendix A for more details about the pretraining dataset.

\noindent\textbf{Training Parameters.}
We utilize the dual-encoder architecture with ResNet50~\citep{he2016deep} as visual encoder and ClinicalBert~\citep{huang2019clinicalbert} as textual encoder, respectively. We train the model with a batch size of $120$/$80$/$25$ for clip-/phase-/video-level, respectively. We sample $4$/$16$/$64$ frames for videos of clip-/phase-/video-level. We use AdamW optimizer \citep{kingma2014adam} with a learning rate of $5e-5$. We train the model with $4$-$80$ GB NVIDIA A$100$ GPUs for $200$ epochs. Temperature parameter $\beta$ for distance function and $\phi$ for DTW-base contrastive loss function $D$ are fixed as $0.1$. Scale factor $\lambda$ is set as $0.01$.

\noindent\textbf{Evaluation Setup.}
We evaluate pretrained models using two setups: Zero-Shot evaluation and Few/Full-shot Linear Probing evaluation. For Zero-Shot, we utilize class text prompts, the same as HecVL~\citep{yuan2024hecvl}, to compute cosine similarities between image embedding and class text embeddings, classifying images based on the shortest distance. In Linear Probing, the pretrained visual encoder remains frozen when we extract features for each image, subsequently training a linear classifier using the SGD optimizer. We consider one shot as one percentage of the videos in the training set because each video contains frames from different classes. Check Appendix B for more details.

\begin{table*}[!htp]
\centering
\caption{Zero-shot phase recognition results. We report Accuracy / F1-Score. PeskaVLP outperforms the other methods across different tasks.}
\label{tab:zero_phase}
\scalebox{0.85}{
    \begin{tabular}{@{}cccccccc@{}}
    \toprule
    Model  & Dataset & Cholec80 & Autolaparo & StrasBypass70 & BernBypass70 & Average  \\
    \midrule

    MIL-NCE~\citep{miech2020end} & Howto100M & 7.8 / 7.3 & 9.9 / 7.9 & 5.6 / 3.1 & 2.4 / 2.1 & 6.4 / 5.1 \\
    \hline

    \multirow{3}{*}{CLIP~\citep{radford2021learning}}  
     & CLIP400M & 30.8 / 13.1 & 17.4 / 9.1 & 16.9 / 5.5 & 14.8 / 4.1 & 19.9 / 8.0 \\

    & Scratch & 29.4 / 10.4 & 15.3 / 10.9 & 6.3 / 3.5 & 4.9 / 2.3 & 14.0 / 6.8 \\

     & SVL & 33.8 / 19.6 & 18.9 / 16.2 & 15.8 / 8.6 & 17.8 / 7.1 & 21.6 / 12.9 \\
    \hline

    SurgVLP~\citep{yuan2023learning} & SVL  & 34.7 / 24.4 & 21.3 / 16.6 & 10.8 / 6.9 & 11.4 / 7.2 & 19.6 / 13.8 \\
    \hline

    HecVL~\citep{yuan2024hecvl} & SVL & 41.7 / 26.3 & 23.3 / 18.9 &  26.9 / 18.3 & 22.8 / 13.6 & 28.7 / 19.3\\

    \hline

    PeskaVLP & SVL  & \textbf{45.1} / \textbf{34.2} & \textbf{26.5} / \textbf{23.6} & \textbf{46.7} / \textbf{28.6} & \textbf{45.7} / \textbf{22.6} & \textbf{41.0} / \textbf{27.1} \\

    \bottomrule
    \end{tabular}
}
\end{table*}

\subsection{Zero-shot Surgical Phase Recognition}

\noindent\textbf{High-quality Surgical Video-language Dataset.} As shown in Tab.~\ref{tab:zero_phase}, our approach achieves a significant performance improvement over the baselines MIL-NCE~\citep{miech2020end} and CLIP~\citep{radford2021learning} pretrained on the natural computer vision datasets, even though our pretraining dataset is $10,000$ times smaller than those. Noted that when the CLIP model is randomly initialized and then trained with SVL, its performance declines compared to initializing from OpenAI. This shows that our surgical video-language pretraining dataset lacks the scale necessary to adequately pretrain a robust video-language model from scratch. ViT~\citep{dosovitskiy2020image,bertasius2021space} architectures are sensitive to initialization and excluded from this work. Further insights into the impact of initialization can be found in Appendix C.

\textbf{Transferability across Surgical Procedures and Centers.} Compared to the HecVL, our method achieves over 12.3\% and 7.8\% improvement in absolute accuracy and f1, thanks to our spatial-temporal $LecNCE$ learning objective across multiple hierarchies. Also, the consistent boost on cholecystectomy~\citep{twinanda2016endonet}, hysterectomy~\citep{wang2022autolaparo}, and gastric bypass~\citep{lavanchy2023challenges} procedures show the generalizable and transferable features of PeskaVLP. Comparing the results of StrasBypass and BernBypass, we find that PeskaVLP can recognize the phases of the same kind of surgery (gastric bypass), even if these surgeries are performed in different centers and follow different procedural routines. More qualitative results can be found in Appendix F.

\subsection{Zero-shot Cross-modal Retrieval}

\begin{table*}[!htp]
\centering
\caption{We present cross-modal retrieval results on the holdout videos, highlighting the best performance in each setting in bold. We additionally include coarser-grained phase-keystep and abstract-video text pairs to assess long-term video and high-level textual understanding.} 

\begin{tabular}{c|ccc|ccc|ccc}
 \hline
& \multicolumn{3}{c|}{Clip-Narration} & \multicolumn{3}{c|}{Phase-Keystep} & \multicolumn{3}{c}{Video-Abstract} \\
\cmidrule(lr){2-10}
method & R@1 & R@5 & R@10 & R@1 & R@5 & R@10 & R@1 & R@5 & R@10 \\
\cmidrule(lr){2-10}
& \multicolumn{9}{c}{Text-to-Image (\%)} \\
\hline
CLIP~\citep{radford2021learning} & 2.9 & 5.2 & 6.7 & 1.7 & 3.2 & 6.3 & 1.2 & 11.7 & 25.8 \\
SurgVLP~\citep{yuan2023learning} & 2.8 & 11.8 & 16.1 & 1.6 & 6.8 & 11.6 & 1.3 & 8.2 & 15.5 \\
HecVL~\citep{yuan2024hecvl} & 2.7 & 11.3 & 17.2 & 3.9 & 13.7 & 21.3 & 28.2 & 74.1 & 82.3 \\
PeskaVLP & \textbf{3.2} & \textbf{13.2} & \textbf{23.3} & \textbf{6.1} & \textbf{21.0} &\textbf{35.4} & \textbf{38.8} & \textbf{75.3} & \textbf{85.9} \\

\midrule
& \multicolumn{9}{c}{Image-to-Text (\%)} \\
\cmidrule(lr){2-10}
CLIP~\citep{radford2021learning} & 1.8 & 3.9 & 6.0 & 0.3 & 1.2 & 2.7 & 0 & 7.0 & 16.4 \\
SurgVLP~\citep{yuan2023learning} & 1.3 & 8.6 & 13.5 & 1.0 & 4.1 & 7.3 & 1.3 & 8.6 & 14.6 \\
HecVL~\citep{yuan2024hecvl} & 2.1 & 9.0 & 16.2 & 1.9 & 8.3 & 14.8 & 21.2 & 65.9 & 71.8 \\
PeskaVLP & \textbf{2.4} & \textbf{13.1} & \textbf{21.3} & \textbf{3.4} & \textbf{14.9} & \textbf{24.8} & \textbf{38.8} & \textbf{75.3} & \textbf{81.1} \\
\hline
\end{tabular}
\label{tab:txt-img}
\end{table*}

In our study, we evaluate pretrained models' cross-modal alignment efficacy by conducting both zero-shot text-to-image and image-to-text retrieval tasks in multiple hierarchical levels. We report the Recall@N metric by identifying the retrieved nearest neighbors for each query and then determining whether the corresponding ground truth element is within the top $N$ nearest neighbors, where $N \in \{1, 5, 10\}$. Tab.~\ref{tab:txt-img} shows that our PeskaVLP achieves superior performance due to the procedure-aware learning objective in hierarchical pretraining. Particularly, the hierarchical pretraining scheme significantly boosts the cross-modal retrieval at the coarse-grained video-text pairs, comprehending the relationship between long video segments and high-level sentences with surgical terms.

\subsection{Few-/Full-shot Linear Probing}

\begin{table*}[!htp]

\centering
\caption{Linear-probing evaluation results. V: supervision is from visual frames. L: supervision is from natural languages. VL: supervision is from both visual and language entities.}
\label{tab:linear_eval}
    \scalebox{0.95}{
    \begin{tabular}{@{}cccccccc@{}}
    \toprule
    Model & Dataset & \%shot & Cholec80 & Autolaparo & StrasBypass70 & BernBypass70 \\
    \midrule

    \multirow{2}{*}{ImageNet} & \multirow{2}{*}{ImageNet (V)} & 100 & 66.4 / 54.9 & 57.5 / 44.9 & 66.2 / 53.6 & 64.7 / 31.6 \\
    && 10 & 57.4 / 42.3 & 44.9 / 30.4 & 53.3 / 42.1 & 53.3 / 25.6 \\
    \hline

    \multirow{2}{*}{Moco~\citep{he2020momentum}} & \multirow{2}{*}{SVL (V)} & 100 & 68.2 / 55.8 & 59.5 / 48.4 & 71.6 / 58.1 & 69.6 / 36.5 \\
    && 10 & 57.6 / 43.5 & 49.9 / 34.6 & 63.1 / 49.3 & 59.1 / 29.9 \\
    \hline

    \multirow{2}{*}{Moco~\citep{he2020momentum}} & \multirow{2}{*}{Cholec80 (V)} & 100 & 73.4 / 62.8 & 51.3 / 37.4 & 67.8 / 55.4 & 66.0 / 33.1 \\
    && 10 & 69.6 / 56.9 & 45.4 / 31.7 & 58.1 / 45.2 & 52.7 / 25.7 \\
    \hline

    \multirow{2}{*}{CLIP~\citep{radford2021learning}} & \multirow{2}{*}{NA (L)} & 100 & 64.8 / 50.7 & 58.5 / 46.1 & 65.4 / 50.6 & 64.1 / 33.3 \\
    && 10 & 57.5 / 40.0 & 46.2 / 31.4 & 54.3 / 42.1 & 52.8 / 27.9 \\
    \hline
    
    \multirow{2}{*}{CLIP~\citep{radford2021learning}} & \multirow{2}{*}{SVL (L)} & 100 & 64.9 / 55.0 & 53.1 / 42.1 & 69.1 / 55.7 & 68.2 / 35.2 \\
    && 10 & 58.9 / 42.3 &  45.3 / 35.3 & 58.2 / 45.2 & 56.5 / 29.8 \\
    \hline

    \multirow{2}{*}{SurgVLP~\citep{yuan2023learning}} & \multirow{2}{*}{SVL (L)} & 100 & 63.5 / 50.3 & 54.3 / 41.8 & 65.8 / 50.0 & 66.5 / 34.3 \\
    && 10 & 55.0 / 39.9 & 48.5 / 32.0 & 57.0 / 44.0 & 57.7 / 28.5 \\
    \hline

    \multirow{2}{*}{HecVL~\citep{yuan2024hecvl}} & \multirow{2}{*}{SVL (L)} & 100 & 66.0 / 53.2 & 56.9 / 44.2 & 69.8 / 54.9 & 70.0 / 34.4 \\
    && 10 & 56.1 / 40.3 & 46.9 / 32.1 & 60.2 / 46.8 & 59.3 / 31.2 \\
    \hline

    \multirow{2}{*}{PeskaVLP} & \multirow{2}{*}{SVL (VL)} & 100 & \textbf{69.9} / \textbf{59.8} & \textbf{63.1} / \textbf{49.7} & \textbf{71.4} / \textbf{59.5} & \textbf{71.5} / \textbf{37.4} \\
    && 10 & \textbf{61.9} / \textbf{50.6} & \textbf{53.1} / \textbf{36.8} & \textbf{63.8} / \textbf{50.4} & \textbf{62.9} / \textbf{32.7} \\ 
    \hline
    
    \bottomrule
    \end{tabular}
    }
\end{table*}

\textbf{General Visual Representation for Surgical Scene Understanding.} We present the few- and full-shot linear-probing evaluation in Tab.~\ref{tab:linear_eval}. It shows that the learned visual representation from PeskaVLP provides a general visual representation for surgical scene understanding across surgical procedures. We also find that the Moco pretrained on the frames of SVL dataset (second row of Tab.~\ref{tab:linear_eval}) in a visual self-supervised manner achieves better visual representation than pretraining on a public dataset that only contains one type of surgery, e.g., Cholec80 (third row in Tab.~\ref{tab:linear_eval}). This shows that the cross-procedure surgical pretraining dataset enables better generalizationability.

\textbf{Knowledge Augmentation and Hierarchical Pretraining.} Interestingly, the model pretrained contrastively with short video clips and narrations (SurgVLP) performs worse than Moco (second row in Tab.~\ref{tab:linear_eval}) in linear probing evaluation. This may be because the noisy narrations do not provide accurate natural language supervision for visual representation learning, thus highlighting the importance of visual self-supervision and textual quality. Our model surpasses the prior methods by a large margin, showing the efficacy of our hierarchical knowledge augmentation, which denoises the text and improves textual quality. Also, our proposed $LecNCE$ promotes the visual encoder through additional visual self-supervision and procedural understanding. We present t-SNE visualizations of learned features in Appendix E, which shows that our multi-modal representations exhibit a smaller modality gap, enhancing transferability to vision-and-language downstream tasks~\citep{gu2023can,liang2022mind}.

\subsection{Ablation Studies}
\label{sec:ablation}

\begin{table*}[t!]
\centering
\caption{Ablation study on different modifications. Knowledge: knowledge augmentation applied to the pretraining dataset. P/V: procedure-aware pretraining learning objective at phase and video-level. C: the integration of language and visual self-supervision at clip-level pretraining. We report 10-shot linear probing in this table.}
\label{tab:ablation}
\scalebox{0.90}{
\begin{tabular}{@{}ccc|cc|cc@{}}
\toprule

\multicolumn{2}{c}{LecNCE} & \multirow{2}{*}{Knowledge} & \multicolumn{2}{c}{Zero-shot} & \multicolumn{2}{c}{Linear-probing} \\
P/V & C & &  Cholec80 & Autolaparo & Cholec80 & Autolaparo  \\
\midrule

$\times$  & $\times$ & $\times$ & 41.7 / 26.3 & 23.3 / 18.9 & 56.1 / 40.3 &  46.9 / 32.1 \\
$\times$  & $\times$ & $\checkmark$ & 42.4 / 28.1 & 24.9 / 20.4 & 58.1 / 43.2 & 48.5 / 34.7\\
$\times$  & $\checkmark$ & $\checkmark$ & 43.4 / 30.3 & \textbf{28.3 }/ \textbf{24.5} & 60.4 / 48.6 & \textbf{53.8} / \textbf{39.2}\\
$\checkmark$ & $\checkmark$ & $\checkmark$ & \textbf{45.1} / \textbf{34.2} & 26.5 / 23.6 & \textbf{61.9} / \textbf{50.6} & 53.1 / 36.8\\ 

\hline
& & & StrasBypass70 & BernBypass70 & StrasBypass70 & BernBypass70  \\
\midrule

$\times$  & $\times$ & $\times$ & 26.9 / 18.3 & 22.8 / 13.6 &  60.2 / 46.8 & 59.3 / 31.2\\
$\times$  & $\times$  & $\checkmark$ & 32.3 / 21.2 & 23.8 / 17.5  & 62.6 / 47.7 & 60.3 / 32.3\\
$\times$  & $\checkmark$ & $\checkmark$ & 39.8 / 23.7 & 25.7 / 21.3 & 63.5 / 48.6 & 62.2 / 32.0\\
$\checkmark$ & $\checkmark$ & $\checkmark$ & \textbf{45.1} / \textbf{34.2}& \textbf{26.5} / \textbf{23.6} & \textbf{63.8} / \textbf{50.4} & \textbf{62.9} / \textbf{32.7}\\
\bottomrule
\end{tabular}
}
\end{table*}

\noindent\textbf{Effect of Knowledge Augmentation.} Tab.~\ref{tab:ablation} presents the effect of our proposed LLM-based hierarchical knowledge-aware augmentation strategy, applied to the texts of SVL dataset. The first row of the table corresponds to HecVL~\citep{yuan2024hecvl} pretrained on SVL with only conventional visual augmentations, e.g., blurring and so on, without any knowledge augmentation. The results clearly demonstrate that simple visual augmentation strategies exhibit poor robustness as the texts of SVL are noisy and not diverse enough. Conversely, our knowledge-aware text augmentation consistently improves performance across multiple surgical datasets, highlighting the importance of the textual quality of the surgical video-language pretraining dataset.

\noindent\textbf{Effect of Pretraining Objective.}
Tab.~\ref{tab:ablation} shows the impact of our learning objective for hierarchical surgical video-language pretraining. When we append visual self-supervision to language supervision at the clip-level pretraining, the zero-shot performance is clearly improved. This improvement can be attributed to the added diverse and high-quality supervision. Also, the boost at linear-probing evaluation shows that the combination of language supervision and visual self-supervision leads to a robust visual representation especially with a moderate size of surgical video-language dataset, e.g., SVL. Table~\ref{tab:ablation} also highlights that the inclusion of $LecNCE$ with procedure understanding consistently improves performance across most downstream datasets, leading to enhanced accuracy in both zero-shot and linear-probing. However, performance on the AutoLaparo degrades with this modification. This may be due to challenging or less routined surgical procedures in the dataset.

\section{Conclusion}

We have introduced a surgical video-language pretraining method for long-term surgical lecture videos and their hierarchical paired texts. Our proposed knowledge augmentation addresses the hierarchical textual information loss by integrating the large language model's internal surgical knowledge. Also, we propose a novel spatial-temporal pretraining objective for video-text pairs of different hierarchies, which addresses the lack of supervision signals problem in a small surgical vision-language dataset. The proposed $LecNCE$ also addresses the procedural awareness problem, benefiting the long-term cross-modal understanding. The experiments show that our proposed PeskaVLP achieves the state-of-the-art generalized zero-shot ability and visual representation learning that can serve as a general initialization for many surgical scene understanding tasks.

\section*{Acknowledgements}
This work has received funding from the European Union (ERC, CompSURG, 101088553). Views and opinions expressed are however those of the authors only and do not necessarily reflect those of the European Union or the European Research Council. Neither the European Union nor the granting authority can be held responsible for them. This work was also partially supported by French state funds managed by the ANR under Grants ANR-20-CHIA-0029-01 and ANR-10-IAHU-02. This work was granted access to the HPC resources of IDRIS under the allocations AD011013704R1, AD011011631R2, and AD011011631R3 made by GENCI. The authors would like to acknowledge the High Performance Computing Center of the University of Strasbourg for supporting this work by providing scientific support and access to computing resources. Part of the computing resources were funded by the Equipex Equip@Meso project (Programme Investissements d'Avenir) and the CPER Alsacalcul/Big Data.

{\small
\bibliographystyle{plain}
\bibliography{main}
}

\newpage

\appendix

\section{Pretraining Dataset}

\subsection{Videos}

We start with the videos that are used for surgical vision-language pretraining in~\citep{yuan2023learning}. In total, there are $1,326$ surgical lecture videos. These videos are transcribed by AWS~\citep{AWS} and Whisper~\citep{radford2022robust} audio speech recognition (ASR) to obtain the corresponding narration texts. Furthermore, we curate the videos' metadata from the online platforms to obtain the extra keystep and abstract texts. In the phase- and video-level pretraining, we need parent- and child-level text correspondences, e.g., keystep and its corresponding narration texts, to perform procedure understanding. Therefore, we filter out the videos that do not have parent-child correspondences. In total, we have $1,007$ and $920$ videos for phase- and video-level pretraining, respectively.

\subsection{Misspelling Error}

As the narration texts are generated from the audio using the ASR system, they usually contain many misspelling errors and fragment sentences. Therefore, we apply multiple preprocessing steps to clean the narration texts.

We first built the vocabulary based on the textbook, surgical category labels, and definition words. Specifically, we refer to the academic papers, which define the surgical phases, to curate a list of definition words and build a vocabulary that contains the words of interest. We also parse and merge the words from the textbook. In total, we obtain a vocabulary of the size of $51,640$ words. Then, we use the built vocabulary along with the spell-checking algorithm~\footnote{https://github.com/barrust/pyspellchecker/} to correct the misspelling errors in narration texts. The algorithm utilizes Levenshtein Distance to identify words within 2 edit distances from the original. It then cross-references these permutations (insertions, deletions, replacements, and transpositions) with a word frequency list, prioritizing words with higher occurrence frequencies as potential correct results.

\section{Evaluation Setup}

We provide a detailed description of the downstream tasks and their settings that we apply in the experiment.

\textbf{Surgical Phase Recognition. } Surgical phase recognition is a proxy task to test the model's surgical scene understanding ability. It aims to classify the frame of surgical video into predefined classes (phases), requiring the model to understand the instrument and anatomy's presence and their interactions by extracting visual patterns from the surgical scene image. In this work, we ignore temporal modeling in surgical phase recognition as we focus on multi-modal representation learning. We consider phase recognition as a frame-wise image classification problem. In the surgical phase recognition task, we evaluate the model's performance based on the publicly available datasets, including Cholec80~\citep{twinanda2016endonet}, AutoLaparo~\citep{wang2022autolaparo} and MultiBypass~\citep{lavanchy2023challenges}.

\begin{table*}[]
	\centering
	\caption{\small{Manually designed prompts for the class names to recognize the surgical phase in Cholec80 dataset. We decompose high-level phase definitions into a few basic concepts to form the text prompts.}}
	\scalebox{0.96}{
		\begin{tabular}{ll}
			\toprule
			\textbf{{Phase Labels}} & \textbf{Prompts} \\
			\midrule
			\emph{Preparation}    & \begin{tabular}[c]{@{}l@{}} In preparation phase I insert trocars to patient abdomen cavity \end{tabular}\\
            \hline
			\emph{CalotTriangleDissection}    & \begin{tabular}[c]{@{}l@{}} In calot triangle dissection phase I use grasper to hold \\ gallbladder  and use hook to expose the hepatic triangle area \\ and cystic duct and cystic artery \end{tabular} \\
            \hline
            \emph{ClippingCutting}    & \begin{tabular}[c]{@{}l@{}} In clip and cut phase I use clipper to clip the cystic duct and \\ artery then use scissor to cut them\end{tabular} \\
            \hline
            \emph{GallbladderDissection}    & \begin{tabular}[c]{@{}l@{}} In dissection phase I use the hook to dissect the connective tissue \\ between gallbladder and liver\end{tabular} \\
            \hline
            \emph{GallbladderPacking}    & \begin{tabular}[c]{@{}l@{}} In packaging phase I put the gallbladder \ into the specimen bag \end{tabular}\\   
            \hline
            \emph{CleaningCoagulation}    & \begin{tabular}[c]{@{}l@{}} In clean and coagulation phase I use suction and irrigation to \\ clear the surgical field and coagulate bleeding vessels \end{tabular} \\      
            \hline
            \emph{GallbladderRetraction}    & \begin{tabular}[c]{@{}l@{}} In retraction phase I grasp the specimen bag and remove \\ it from trocar \end{tabular}\\\bottomrule\hline
		\end{tabular}
	}
	\label{table:prompt1}
	\vspace{-1mm}
\end{table*}

\begin{itemize}
    \item \textbf{Zero-shot Evaluation. }As the surgical phase labels are high-level definitions that can be decomposed into a few basic concepts, we manually construct the contextual prompts for phase labels, as shown in Tab.~\ref{table:prompt1}, Tab.~\ref{table:prompt2} and Tab.~\ref{table:prompt3}. Our constructed prompts for the class names are built with the help of clinician's comments, considering the involved surgical instruments, anatomies, and events involved in a given surgical phase.  
    \item \textbf{Linear-probing Evaluation. } For linear-probing evaluation on the surgical phase recognition downstream datasets, we keep the visual encoder frozen and train a linear classifier on the extracted features. We do not apply any image augmentation during the training. The learning rate is scaled linearly based on the actual batch size. The model is optimized using SGD optimizer with the learning rate as $0.001$ and weight decay parameter as $0.0005$. We train the model for $40$ epochs. We fit the model on the training and validation sets and report the performance on the separate test set. For the few-shot linear-probing evaluation, we adopt an N-way K-shot approach with a slight modification to accommodate the nature of surgical videos, which contain frames from different classes. Specifically, we select 10\% of the video from the training set. This ensures that data leakage is prevented and that the number of samples per class remains similar.
\end{itemize}

\textbf{Cross-modal Retrieval. } Cross-modal retrieval includes text-based video retrieval and video-based text retrieval. Here, we conduct the cross-modal retrieval at three hierarchical levels.  We collect $537$ clip-narration (clip-level) video-text pairs, $746$ phase-keystep (phase-level) video-text pairs, and $86$ video-abstract (video-level) video-text pairs from hold-out testing videos of SVL~\citep{yuan2023learning}. There are more phase-keystep than clip-narration video-text pairs because some testing videos do not have cleaned narrations and we filter them out. For video embedding generation, we sample multiple frames fro
m the video and average pool their image embeddings. We temporally sample $10$ frames for clip-/phase-/video-level videos. We conduct the zero-shot evaluation for the cross-modal retrieval task.

\begin{table*}[!t]
	\centering
	\caption{\small{Manually designed prompts for the class names to recognize the surgical phase in AutoLaparo dataset.}}
	\scalebox{0.96}{
		\begin{tabular}{ll}
			\toprule
			\textbf{{Phase Labels}} & \textbf{Prompts} \\
			\midrule
			\emph{Preparation}    & \begin{tabular}[c]{@{}l@{}} I use grasper to grasp and explore the field \end{tabular}\\
            \hline
			\emph{Dividing Ligament and Peritoneum}    & \begin{tabular}[c]{@{}l@{}} I divide ligament and peritoneum \end{tabular} \\
            \hline
            \emph{Dividing Uterine Vessels and Ligament}    & \begin{tabular}[c]{@{}l@{}}  I divide uterine vessels and ligament \end{tabular} \\
            \hline
            \emph{Transecting the Vagina}    & \begin{tabular}[c]{@{}l@{}} I use the dissecting hook to transect the vagina \end{tabular}  \\
            \hline
            \emph{Specimen Removal}    & \begin{tabular}[c]{@{}l@{}} I remove the specimen bag and uterus  \end{tabular}\\   
            \hline
            \emph{Suturing}    & \begin{tabular}[c]{@{}l@{}} I suture the tissue \end{tabular} \\      
            \hline
            \emph{Washing}    & \begin{tabular}[c]{@{}l@{}} Washing \end{tabular}\\\bottomrule\hline
		\end{tabular}
	}
	\label{table:prompt2}
	\vspace{-1mm}
\end{table*}

\begin{table*}[!t]
	\centering
	\caption{\small{Manually designed prompts for the class names to recognize the surgical phase in gastric bypass dataset. We use the same prompts for both StrasBypass70 and BernBypass70. We exclude the ``other'' class as its definition is ambiguous.}}
	\scalebox{0.98}{
		\begin{tabular}{ll}
			\toprule
			\textbf{{Phase Labels}} & \textbf{Prompts} \\
			\midrule
			\emph{Preparation}    & \begin{tabular}[c]{@{}l@{}} In preparation phase I insert trocars to the abdominal cavity  \\ and expose of the operating field
 \end{tabular}\\
            \hline
			\emph{Gastric pouch creation}    & \begin{tabular}[c]{@{}l@{}} I cut the fat tissue and open retrogastric window at stomach \end{tabular} \\
            \hline
            \emph{Omentum division}    & \begin{tabular}[c]{@{}l@{}} I grasp and lift the omentum and divide it\end{tabular} \\
            \hline
            \emph{Gastrojejunal anastomosis}    & \begin{tabular}[c]{@{}l@{}} I see the proximal jejunum and determine the length of \\the biliary limb. I open the distal jejunum and create the \\ gastrojejunostomy using a stapler. I reinforcement of the \\ gastrojejunostomy with an additional suture.\end{tabular} \\
            \hline
            \emph{Anastomosis test}    & \begin{tabular}[c]{@{}l@{}} I place the retractor and move the gastric tube and detect \\ any leakage of the gastrojejunostomy\end{tabular}\\   
            \hline
            \emph{Jejunal separation}    & \begin{tabular}[c]{@{}l@{}} I open the mesentery to facilitate the introduction of the \\ stapler and transect the jejunum proximal \end{tabular} \\      
            \hline
            \emph{Petersen space closure}    & \begin{tabular}[c]{@{}l@{}} I expose between the alimentary limb and the transverse \\ colon and close it with sutures \end{tabular} \\
            \hline
            \emph{Jejunojejunal anastomosis}  & \begin{tabular}[c]{@{}l@{}} I expose between the alimentary limb and the transverse \\ colon and close it with sutures \end{tabular} \\
            \hline
            \emph{Mesenteric defect closure}  & \begin{tabular}[c]{@{}l@{}} I expose the mesenteric defect and then close it by stitches \end{tabular} \\
            \hline
            \emph{Cleaning and coagulation}  & \begin{tabular}[c]{@{}l@{}} In clean and coagulation phase I use suction and irrigation\\ to clear the surgical field and coagulate bleeding vessels \end{tabular} \\  
            \hline
            \emph{Disassembling}  & \begin{tabular}[c]{@{}l@{}} I remove the instruments, retractor, ports, and camera \end{tabular} \\  
            \bottomrule\hline
		\end{tabular}
	}
	\label{table:prompt3}
	\vspace{-1mm}
\end{table*}

\section{Architecture \& Initialization}

\begin{table*}[!]
    \centering
    \label{tab:init_arch}
    \scalebox{0.9}{
    \begin{tabular}{cccccccc}
    \toprule[1pt]
    \multirow{2}{*}{\footnotesize\textbf{Backbone}} & \multirow{2}{*}{\footnotesize\textbf{Init.}} &\multicolumn{2}{c}{\footnotesize{Zero-shot}} & \multicolumn{2}{c}{\footnotesize{Linear-probing} (10-shot)} & \multicolumn{2}{c}{\footnotesize{Linear-probing} (full-shot)} \\
    \cmidrule(lr){3-4}\cmidrule(lr){5-6}\cmidrule(lr){7-8}
    & & \footnotesize\textbf{Cholec80}  & \footnotesize\textbf{Autolaparo} & \footnotesize\textbf{Cholec80}  & \footnotesize\textbf{Autolaparo} & \footnotesize\textbf{Cholec80}  & \footnotesize\textbf{Autolaparo} \\
    \midrule
    \multirow{4}{*}{ResNet50} & \footnotesize{Random} & 29.4 / 10.4 & 15.3 / 10.9 & 42.4 / 22.1 & 33.4 / 20.2 & 44.6 / 25.3 & 30.7 / 19.3 \\
    & \footnotesize{ImageNet} & 34.7 / 24.4 & 21.3 / 16.6 & 55.0 / 39.9 & 48.5 / 32.0 & 63.5 / 50.3 & 54.3 / 41.8 \\
    & \footnotesize{CLIP} & 33.8 / 19.6 & 18.9 / 16.2 & 58.9 / 42.3 &  45.3 / 35.3 & 64.9 / 55.0  & 53.1 / 42.1 \\
    \midrule
    \multirow{4}{*}{ViT-B/16} & \footnotesize{Random} & 20.2 / 11.5 & 9.1 / 8.3 & 38.4 / 20.9 & 32.1 / 19.7 & 48.2 / 25.9 & 38.4 / 25.5 \\
    & \footnotesize{ImageNet} & 42.8 / 25.1 & 20.5 / 15.5 & 57.4 / 40.5 & 47.8 / 31.9 & 60.6 / 48.9 & 56.3 / 44.5 \\
    & \footnotesize{Dino} & 35.1 / 19.1 & 13.9 / 9.2 & 54.7 / 39.2 & 47.4 / 31.1 & 64.9 / 51.2 & 54.0 / 42.4\\
    \bottomrule[1.2pt]
    \end{tabular}
    }
    \caption{The experiments show that the initialization largely influences the performance of surgical video-language pretraining.}
\end{table*}

As mentioned before, the current surgical vision-language pretraining dataset lacks the scale necessary to pretrain a robust vision-language model from scratch, therefore a good choice of architecture and initialization is important. In this section, we conduct the experiment and study the effect of different model architectures and initializations, justifying our choice of using ResNet50 architecture with ImageNet initialization as our starting point before the video-language pretraining. 

\begin{itemize}
    \item ResNet50. For ImageNet initialization, we use public IMAGENET1K\_V1 weights from torchvision. Random initialization means that we random initialize the visual encoder before the hierarchical vision-language pretraining. These models' textual encoders are initialized from BioClinicalBert~\citep{huang2019clinicalbert}. For CLIP initialization, we initialize the visual and textual encoder from OpenAI's weight~\citep{radford2021learning}. 
    \item ViT-B/16. For ImageNet initialization, we use weights from the official Google JAX implementation, which is pretrained on ImageNet21k~\citep{ridnik2021imagenet} and then finetune on ImageNet1k~\citep{russakovsky2015imagenet}. We use the public pretrained weights from~\citep{caron2021emerging} for Dino initialization. 

\end{itemize}

In our work, we choose ResNet50 over Vision Transformer (ViT-B/16) due to its superior performance and lower parameter amounts in the context of video-language pretraining for surgical data. Our experiments demonstrated that ResNet50, particularly when initialized with CLIP weights, outperformed ViT-B/16 across various tasks, including zero-shot and linear-probing evaluations on Cholec80 and Autolaparo datasets. Despite the advanced capabilities of vision transformers, their performance heavily depends on large-scale pretraining datasets, which might not always be available or optimal for specialized domains like surgical scenes. Conversely, convolutional neural networks like ResNet50 have shown robust generalization abilities, even when pretrained on natural images, making them more suitable for our specific application. Additionally, the initialization sensitivity observed in ViT-B/16 further justified our preference for ResNet50, ensuring a more reliable and effective starting point for our hierarchical vision-language pretraining.

\section{Dynamic Time Warping}
After achieving the cost matrix $C$ and $\hat{C}$, we perform dynamic time warping (DTW)~\citep{sakoe1978dynamic} to find the minimum cost path to align the frames of video segment $V=\{v_1,...v_T\}$ to the text sequence $B=\{b_1,...b_N\}$ and reversed text sequence $\{b_N,...b_1\}$, respectively, as shown in Algorithm.~\ref{algorithm:dtw}. We follow~\citep{xue2024learning} to process the DTW function into differentiable, enabling the gradient back-propagation. The differentiable loss function is the same as~\citep{hadji2021representation}.

A significant advantage of using DTW is that it does not require additional temporal modules, such as recurrent neural networks or attention mechanisms, to model temporal relationships. This simplification allows us to focus on learning better representations by directly aligning video frames and text sequences based on their semantics.

\begin{algorithm}
    \caption{DTW to align sequences using cost matrix}
    \begin{algorithmic}[1]
        \Procedure{AlignSequences}{$C, V, B$}
            \State Let $T$ be the length of sequence $V$ and $N$ be the length of sequence $B$.
            \State Set $i$ to $T$ and $j$ to $N$.
            \State Initialize $distance$ to $0$.
            \While{$i > 0$ and $j > 0$}
                \State $distance$ = $distance$ + $C[i][j]$
                \If{$i > 1$ and $j > 1$ and $C[i-1][j-1] \leq C[i-1][j]$ \textbf{and} $C[i-1][j-1] \leq C[i][j-1]$}
                    \State $i \gets i - 1$
                    \State $j \gets j - 1$
                \ElsIf{$i > 1$ and $C[i-1][j] \leq C[i][j-1]$}
                    \State $i \gets i - 1$
                \Else
                    \State $j \gets j - 1$
                \EndIf
            \EndWhile
            \State \textbf{return} $distance$.
        \EndProcedure
    \end{algorithmic}
    \label{algorithm:dtw}
\end{algorithm}

\section{Modality Gap}

Modality gap is a geometric phenomenon observed in the embedding space of multi-modal models~\citep{liang2022mind}. This gap illustrates that pretrained multi-modal (vision-language) models create a joint embedding space where different modalities, such as images and text, are kept at a significant distance from each other. During contrastive optimization, this separation created at initialization is maintained to the extent that irrelevant image embeddings can be closer to each other than to their corresponding relevant text embeddings. This spatial disparity in the embedding space hinders the model's ability to effectively align and understand the relationships between visual and textual data, leading to suboptimal performance in tasks requiring integrated multi-modal comprehension. The existence of the modality gap is particularly detrimental when adapting pretrained vision-language models to cross-modal generation tasks, such as image captioning. As highlighted by several studies~\citep{li2023decap,gu2023can}, narrowing modality gap correlates with improved performance in cross-modal tasks. 

\begin{figure}
    \centering
    \includegraphics[width=0.8\textwidth]{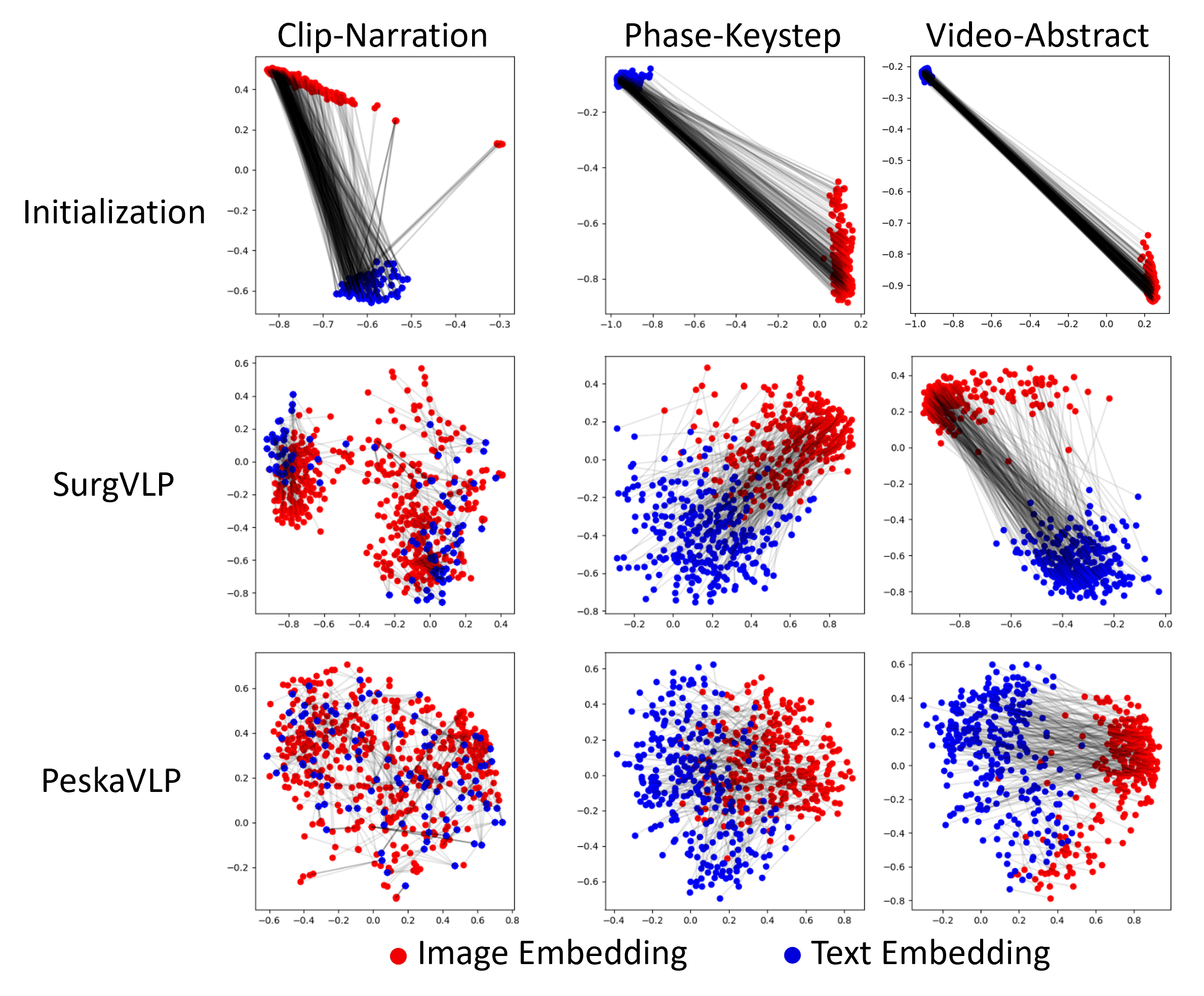}
    \caption{Modality gap visualization in different hierarchical levels. It shows that our model closes the modality gap incurred from the initialization after the hierarchical pretraining.}
    \label{fig:modality_gap}
\end{figure}

As shown in Fig.~\ref{fig:modality_gap}, we visualize the embeddings of videos and their corresponding text descriptions at three hierarchical levels: clip-narration, phase-keystep, and video-abstract. Our proposed model demonstrates a significant reduction in the modality gap compared to the SurgVLP model. This alignment across different hierarchical levels ensures a more comprehensive and cohesive understanding of the multi-modal data, leading to superior performance in tasks like image captioning and other vision-language applications.

\section{Surgical Phase Recognition Results}
We demonstrate the zero-shot surgical phase recognition to reflect the surgical scene understanding ability of our pretrained model. Our model can identify surgical phases of different types of surgical procedures without any finetuning. Both success and failure examples are shown.

\begin{figure}[!]
    \centering
    \includegraphics[width=\textwidth]{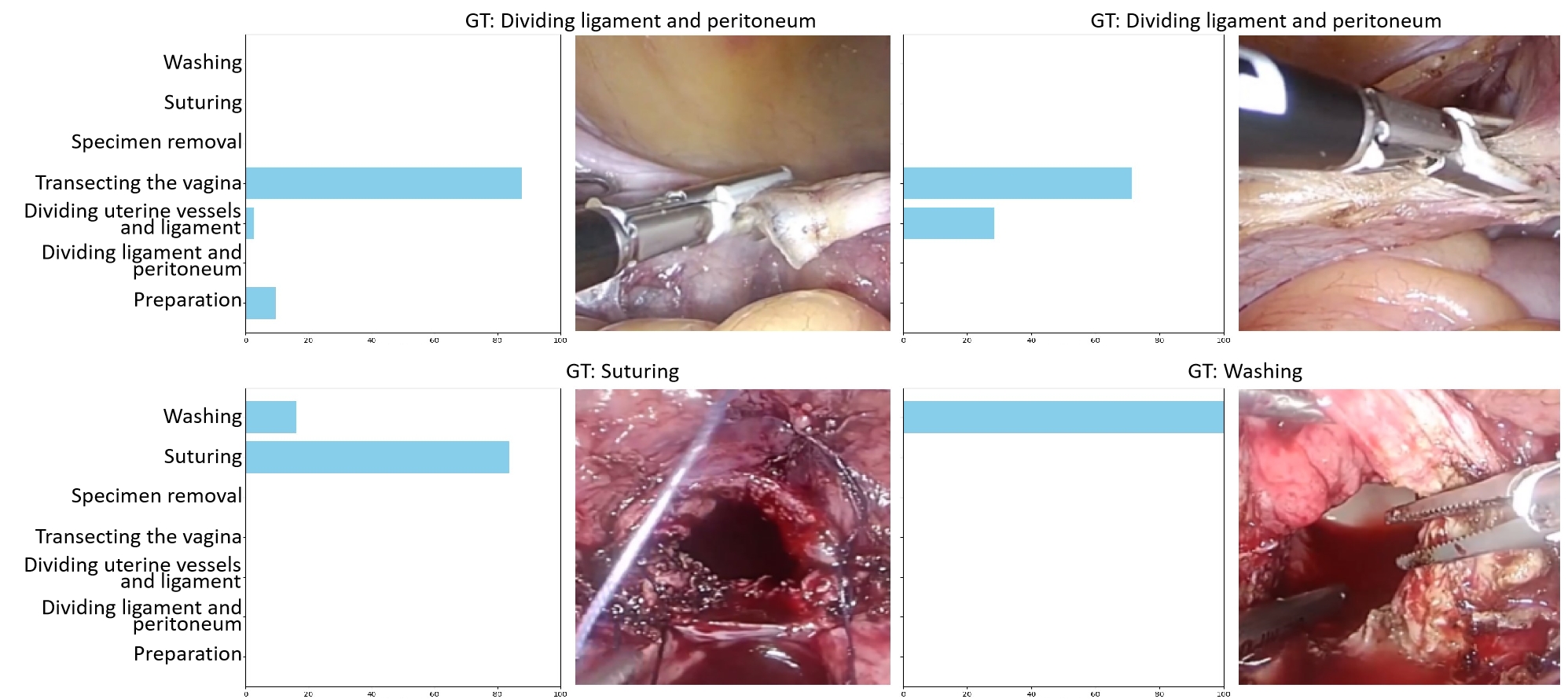}
    \caption{Qualitative surgical phase recognition results on hysterectomy. The y-axis is the class names. The x-axis is the probability of each class. The bottom right image shows that the pretrained model understands the blood fluid.}
    \label{fig:qualitative_autolapa}
\end{figure}

\begin{figure}[!]
    \centering
    \includegraphics[width=\textwidth]{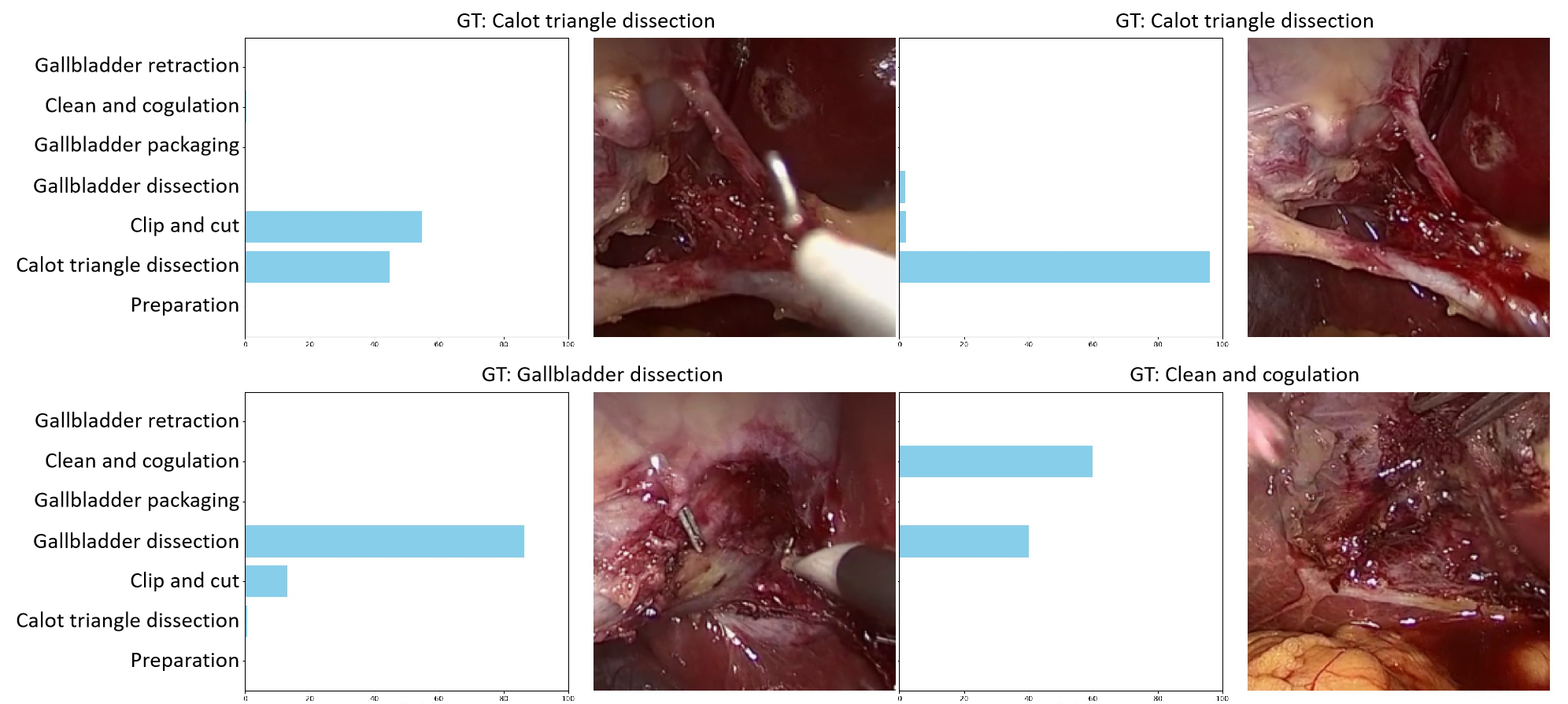}
    \caption{Qualitative surgical phase recognition results on cholecystectomy. The y-axis is the class names. The x-axis is the probability of each class. We find that the pretrained model is triggered by the instrument occurrence, such as hook in the second row.}
    \label{fig:qualitative_c80}
\end{figure}

\textbf{Surgical Term Understanding.} In Fig.~\ref{fig:qualitative_autolapa}, we show that the pretrained model excels at identifying the ``washing'' phase in surgical procedures, demonstrating its capability to accurately recognize high-level surgical activities. This proficiency enhances surgical assistance systems, improving real-time analysis and decision-making in operating rooms.

\textbf{Instrument Identification.} In Fig.~\ref{fig:qualitative_c80}, we demonstrate how the visual embedding is significantly influenced by the presence of surgical instruments. Specifically, in the first row, the semantic meaning of the image changes from "calot triangle dissection" to "clip and cut" due to the appearance of a hook, even though the other anatomical features remain similar.

\section{Limitations}
As the pretraining process at clip-level requires additional supervision signals, i.e., visual self-supervision, the memory and computation overhead increase compared to the vanilla HecVL pretraining. Also, during the phase- and video-level pretraining, the process of dynamic time warping can be time-consuming because it is based on dynamic programming, slowing down the pretraining iteration when handling longer-term surgical videos. Additionally, the knowledge augmentation on keystep and abstract texts need to be modified to fit the other video-language pretraining datasets~\citep{ashutosh2023hiervl,zhang2018cross} as their hierarchical paired texts are annotated manually. Instead, our knowledge augmentation is more suitable for videos in the wild from online platforms. 
To address these limitations, future work could focus on developing a general textual augmentation strategy using the LLM's internal knowledge, adapting to the instructional videos that miss keystep and abstract text descriptions. Furthermore, techniques for decentralizing the video-language pretraining could be explored, aiming to pretrain with multi-centric vision-language samples while preserving privacy using the federated learning strategy. This could address the scaling problem in surgical vision-language pretraining and improve the generalizationability across the centers.

\section{Knowledge Augmentation}

\begin{figure}[!]
    \centering
    \includegraphics[width=1\textwidth]{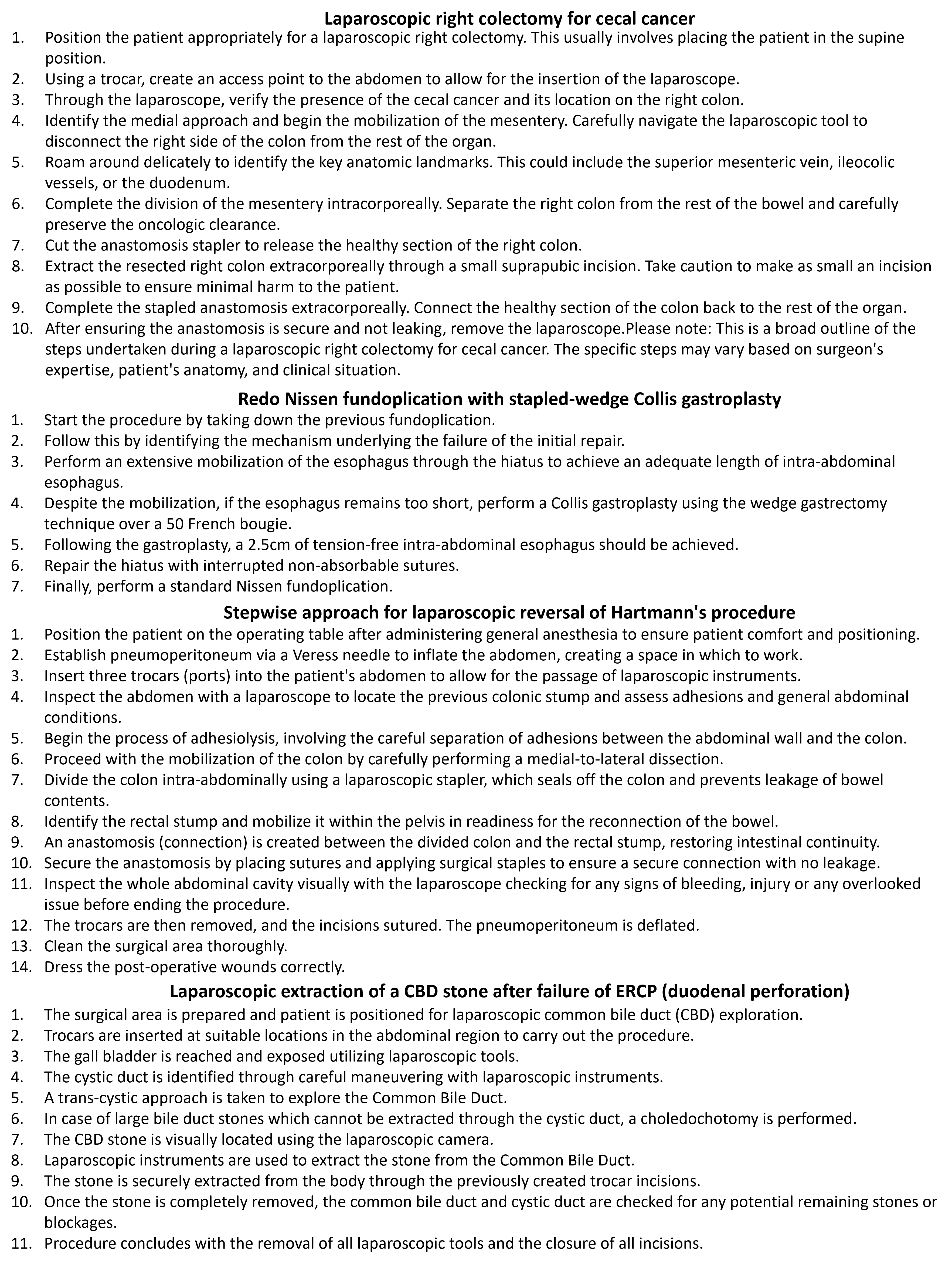}
    \caption{Example of surgical step knowledge base based on the large language models.}
    \label{fig:build_kb}
\end{figure}

\textbf{Build Surgical Knowledge Base. } In Fig.~\ref{fig:build_kb}, we show that the internal surgical knowledge of large language models can be elicited to build the external knowledge base.

\begin{figure}[!]
    \centering
    \includegraphics[width=1\textwidth]{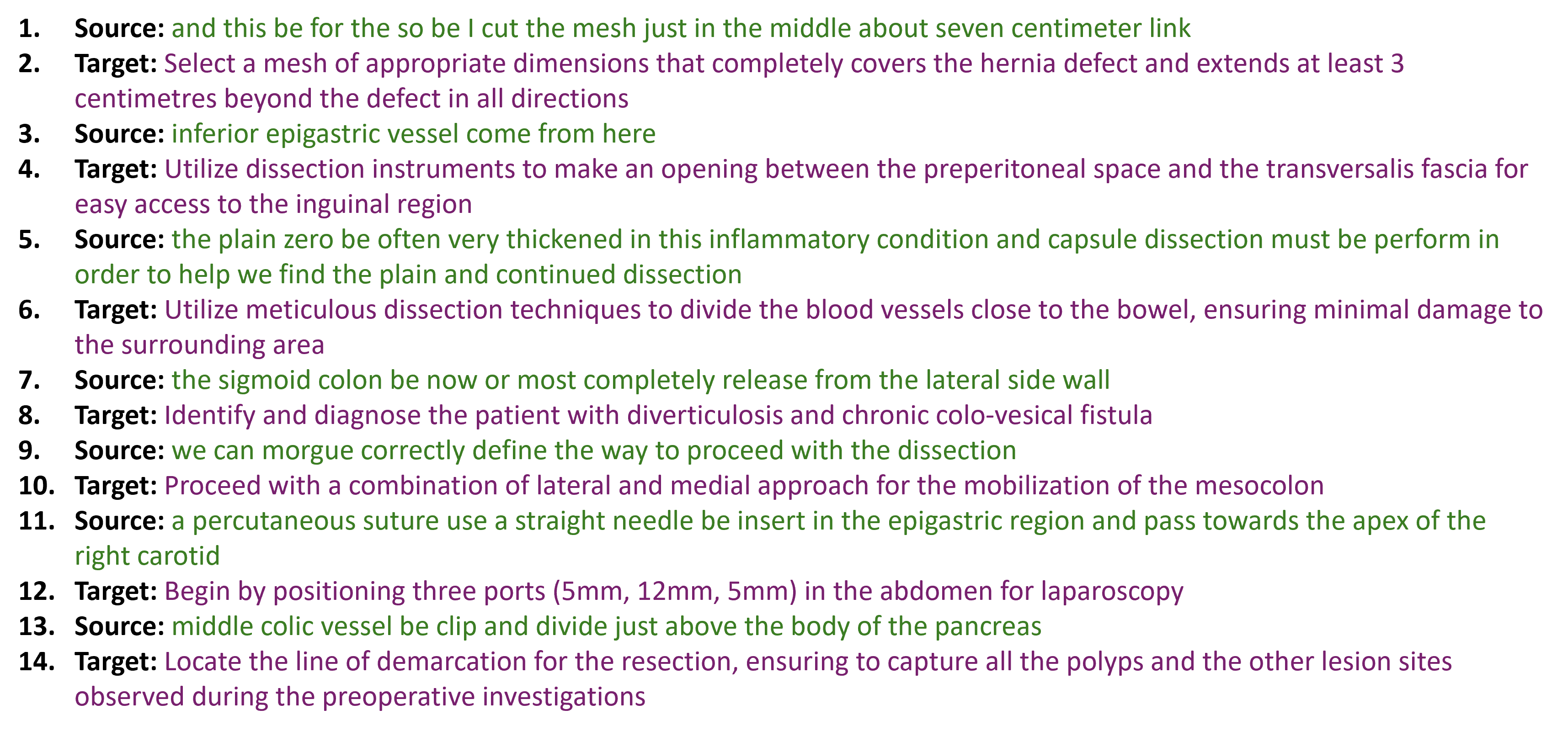}
    \caption{Knowledge augmentation on the narration texts.}
    \label{fig:narration_rewrite}
\end{figure}

\begin{figure}[!]
    \centering
    \includegraphics[width=1\textwidth]{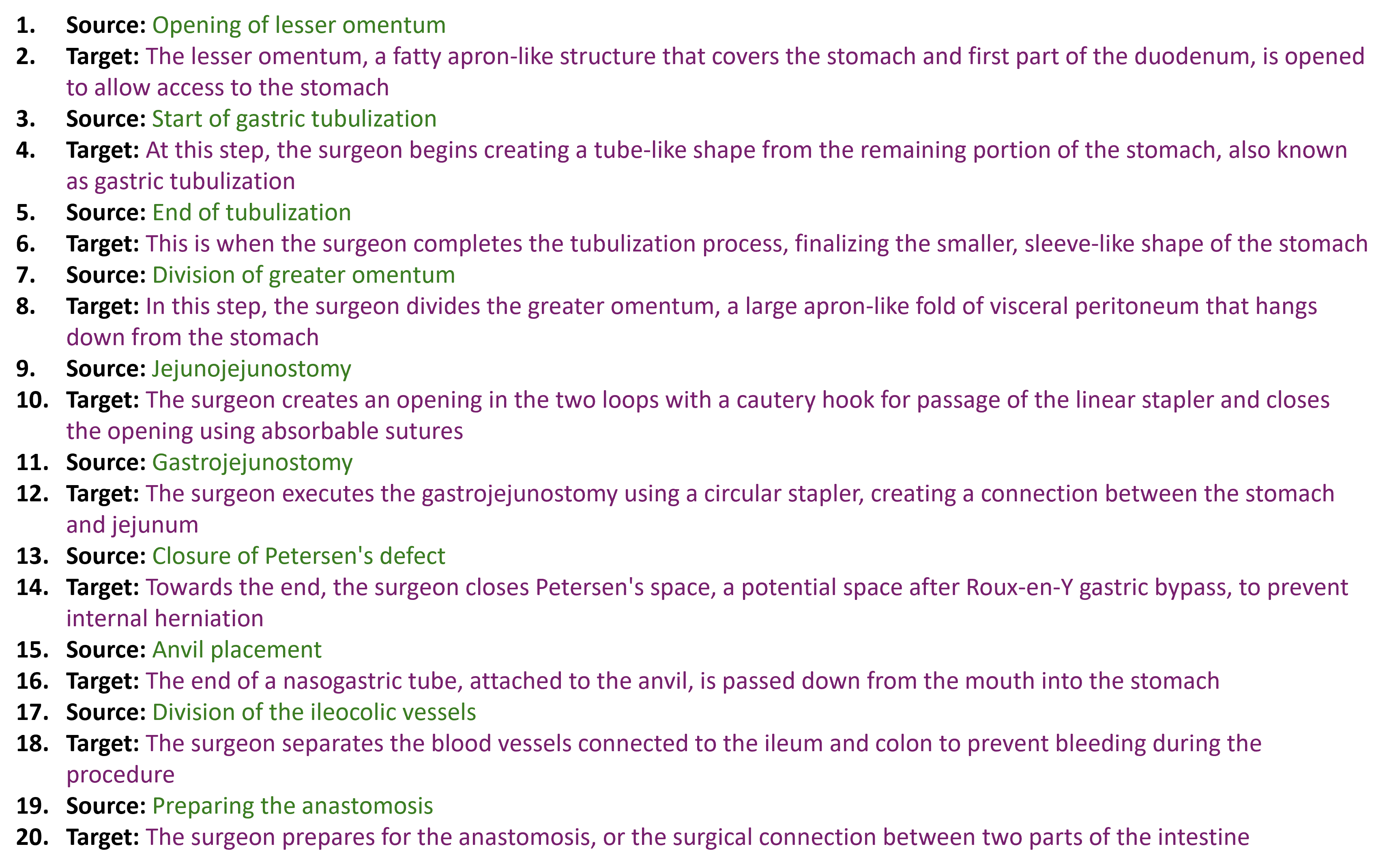}
    \caption{Knowledge augmentation on the keystep texts.}
    \label{fig:keystep_rewrite}
\end{figure}

\begin{figure}[!]
    \centering
    \includegraphics[width=1\textwidth]{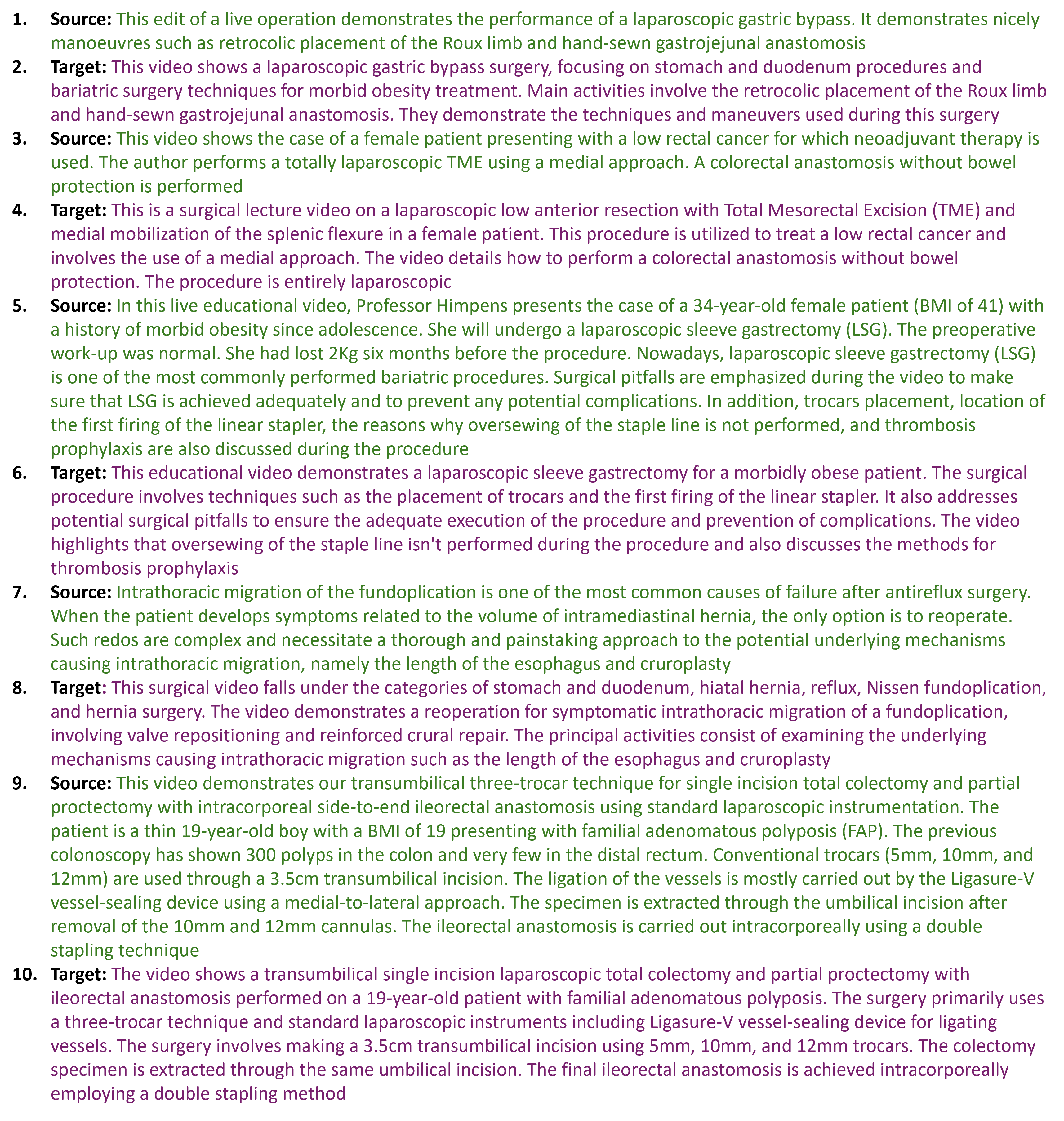}
    \caption{Knowledge augmentation on the abstract texts.}
    \label{fig:abstract_rewrite}
\end{figure}

\textbf{Build Surgical Knowledge Base. } In Fig.~\ref{fig:narration_rewrite}, Fig.~\ref{fig:keystep_rewrite} and Fig.~\ref{fig:abstract_rewrite}, we show that the knowledge of large language model can be used to enrich the semantics of the hierarchical texts, i.e., narrations, keysteps, and abstracts. Notably, it can explain high-level keystep words into descriptive sentences, enhancing textual diversity and preventing overfitting.

\end{document}